\documentclass[10pt,twocolumn,letterpaper]{article}

\usepackage[pagenumbers]{cvpr} % To force page numbers, e.g. for an arXiv version

\definecolor{cvprblue}{rgb}{0.21,0.49,0.74}
\usepackage[pagebackref,breaklinks,colorlinks,allcolors=cvprblue]{hyperref}

\def\paperID{*****} % *** Enter the Paper ID here
\def\confName{CVPR}
\def\confYear{2025}

\title{Taming VR Teleoperation and Learning from Demonstration for Multi-Task Bimanual Table Service Manipulation}

\author{
\textbf{MilkDragon Team}\\
Weize Li$^{1}$\thanks{Corresponding author.} \quad Zhengxiao Han$^{2}$\quad Lixin Xu$^{1,3}$\quad Xiangyu Chen$^{1,4}$\\
Harrison Bounds$^{2}$\quad Chenrui Zhang$^{1,5}$\quad Yifan Xu$^{6}$\\
$^{1}$AIR, Tsinghua University\quad $^{2}$Northwestern University\quad $^{3}$Georgia Institute of Technology\\
$^{4}$HKUST (GZ)\quad $^{5}$NUS (Suzhou) Research Institute\quad $^{6}$University of Michigan\\
\texttt{liweize0224@gmail.com}
}

\begin{document}
\maketitle
\begin{abstract}
This technical report presents the champion solution of the Table Service Track in the ICRA 2025 What Bimanuals Can Do (WBCD) competition.  
We tackled a series of demanding tasks under strict requirements for speed, precision, and reliability: unfolding a tablecloth (deformable-object manipulation), placing a pizza into the container (pick-and-place), and opening and closing a food container with the lid.  
Our solution combines \textbf{VR-based teleoperation} and \textbf{Learning from Demonstrations (LfD)} to balance robustness and autonomy.  
Most subtasks were executed through high-fidelity remote teleoperation, while the pizza placement was handled by an ACT-based policy trained from 100 in-person teleoperated demonstrations with randomized initial configurations.
By carefully integrating scoring rules, task characteristics, and current technical capabilities, our approach achieved both high efficiency and reliability, ultimately securing the first place in the competition.

\end{abstract}
\section{Introduction}
\label{sec:intro}
In the past decade, bimanual manipulation and learning from demonstration (LfD) have emerged as promising frameworks that enable robots to acquire complex skills from human operators. Two-handed coordination is fundamental to many real-world tasks—from industrial assembly to household service—and has therefore drawn increasing attention from both academia and industry. Leveraging advances in robot hardware, wearable interfaces, and policy learning algorithms, platforms such as ALOHA~\cite{zhao2023learning}, DexCap~\cite{wang2024dexcap}, Mobile ALOHA~\cite{fu2024mobile}, and Open-TeleVision~\cite{cheng2024open} have showcased convincing demonstrations of human-to-robot skill transfer in increasingly dexterous scenarios. On the algorithmic side, progress has been made at two levels: generalist vision-language-action (VLA) policies such as $\pi$-0~\cite{black2024pi_0} and RDT-1B~\cite{liu2024rdt} that aim for broad task coverage, and specialist policies such as Diffusion Policy~\cite{chi2023diffusion}, BC~\cite{torabi2018behavioral}, and ACT~\cite{zhao2023learning} that focus on stable and efficient learning for specific manipulation tasks.

Despite these advances, most research is still evaluated on self-designed tasks, often neglecting crucial industry metrics such as operational speed, reliability, and autonomy. The ICRA 2025 WBCD Challenge – Table Service Track directly addresses this gap by introducing a series of tasks that combine deformable manipulation, precision operations, and coordinated actions, which pose challenges that cannot be solved by a single paradigm alone.

In this technical report, we present the MilkDragon team’s champion solution, which integrates VR-based teleoperation with imitation learning. Our design philosophy is to leverage teleoperation for robustness in variable or force-sensitive subtasks, while using an ACT-based policy to automate reliable pick-and-place. This hybrid approach allows us to balance efficiency and reliability under strict competition constraints, ultimately securing first place in the challenge. Our results further highlight the potential of hybrid pipelines to bridge research prototypes and real-world service robotics applications.

\begin{figure*}
  \centering
  \includegraphics[width=1 \textwidth]{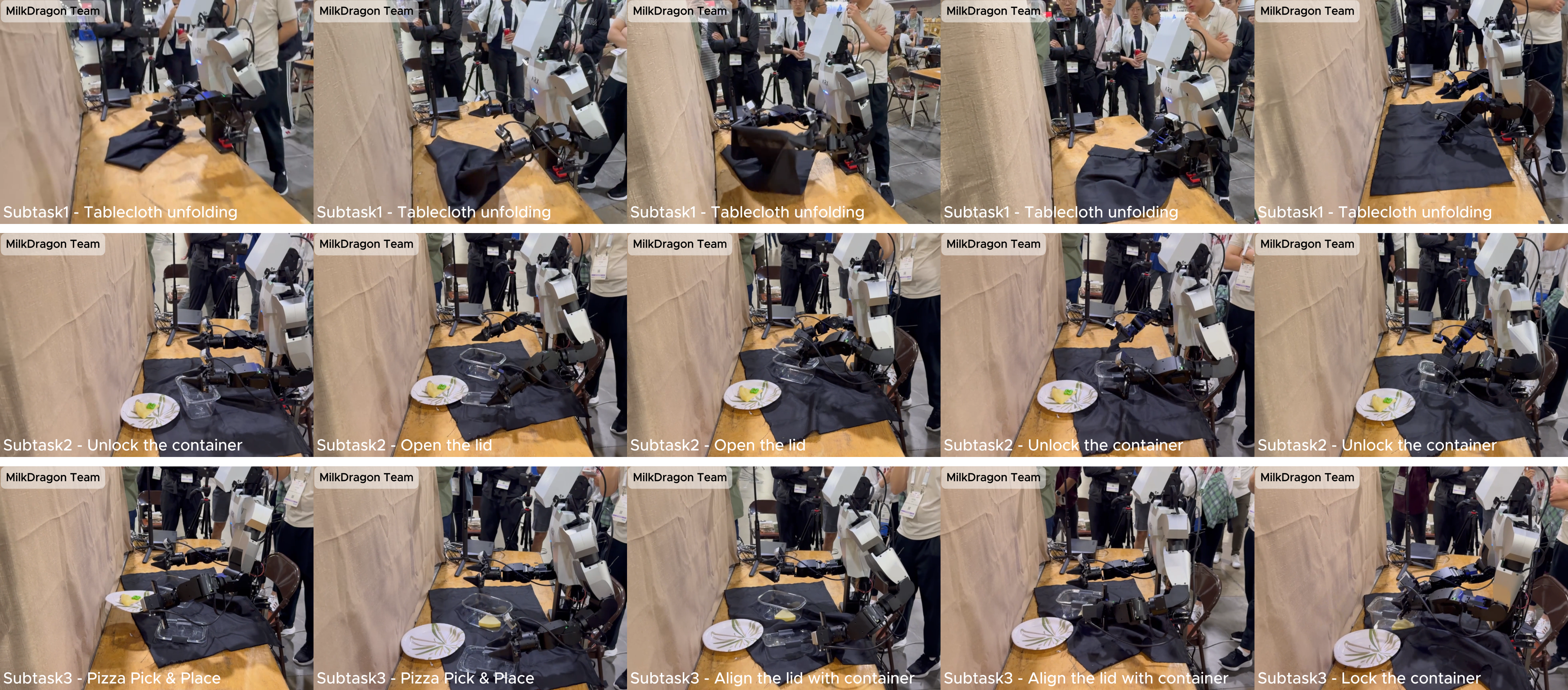}
  % \vspace{-20pt}
  \caption{Overview of our team's bimanual manipulation task flow in the ICRA 2025 WBCD Challenge - Table Service Track.
  }
  \label{fig:teaser}
\end{figure*}
\section{Challenge Setup: Table Services Track}
\subsection{Hardware}
\paragraph{Environment Setup.} We were allocated a 10 ft × 15 ft workspace equipped with one table and one chair. The table measured approximately 8 ft × 30 in × 29 in (L×W×H). Power outlets (120V, 1500W) were available at each station.

\paragraph{Robot Platform.} In the competition, we utilized the competition edition of the X7s robot provided by hardware sponsor ARX Robotics. The X7s is a humanoid data acquisition platform featuring a head with 2 DoFs, and two robotic arms each with 7 DoFs plus 1 DoF at the end-effector gripper. Notably, due to shipping constraints, the competition edition of the X7s does not include the mobile base or lifting mechanism. The robot's upper body is fixed to the table using 3 G-clamps.

\noindent \subsection{Task Description and Scoring Criteria} In the Table Services Track, each team has a 30-minute time window during which we are required to use the props provided by the organizers as shown in Fig. 2 and complete as many task rounds as possible in the prescribed order. Each subtask is worth up to 5 points, for a maximum of 15 points per round. The task descriptions are as follows:

\begin{itemize}
    \item \textbf{Subtask 1: Tablecloth Unfolding}
    \begin{itemize}
        \item Fully unfold a pre-folded tablecloth (typically folded 4--5 times) into a flat, rectangular shape: \textbf{5 points}
    \end{itemize}
    
    \item \textbf{Subtask 2: Opening the Food Container}
    \begin{itemize}
        \item Unlock two sides of the container: \textbf{3 points}
        \item Remove the lid: \textbf{2 points}
    \end{itemize}

    \item \textbf{Subtask 3: Packing Pizza}
    \begin{itemize}
        \item Place the pizza from plate into the container: \textbf{1 point}
        \item Align the lid precisely with the container: \textbf{2 points}
        \item Lock two sides of the container: \textbf{2 points}
    \end{itemize}
\end{itemize}

\noindent For the scoring criteria, each subtask score was determined by three factors:

\begin{itemize}
    \item $\alpha$: the \textbf{operation coefficient}, reflecting autonomy. 
    It takes values $\{0.5,1,4\}$ for in-person teleoperation, remote teleoperation, and autonomous policy, respectively.
    
    \item $\beta$: the \textbf{completion time} (in seconds). 
    Faster executions correspond to smaller $\beta$, and thus yield higher efficiency.
    
    \item $s$: the \textbf{base score} of the subtask, defined by successful completion 
    (up to 5 points per subtask).
\end{itemize}
The final score for a subtask is computed as:
\[
\text{Score}_{\text{subtask}} = \frac{s}{\beta} \times \alpha.
\]
The round score is then:
\[
S_{\text{round}} = \sum_{k=1}^{3} \frac{s_k}{\beta_k} \, \alpha_k,
\]
and the total competition score is obtained by summing across all rounds:
\[
S_{\text{total}} = \sum_{\text{rounds}} S_{\text{round}}.
\]
This formulation ensures that the scoring captures three key aspects simultaneously: 
\emph{task accuracy} ($s$), \emph{autonomy level} ($\alpha$) and \emph{execution efficiency} ($\beta$ in the denominator). 
As a result, teams were incentivized to complete tasks correctly, as quickly as possible, and with higher levels of autonomy.

% \url{https://wbcdcompetition.github.io/}.

\begin{figure}[h]
    \centering
    \includegraphics[width=\columnwidth]{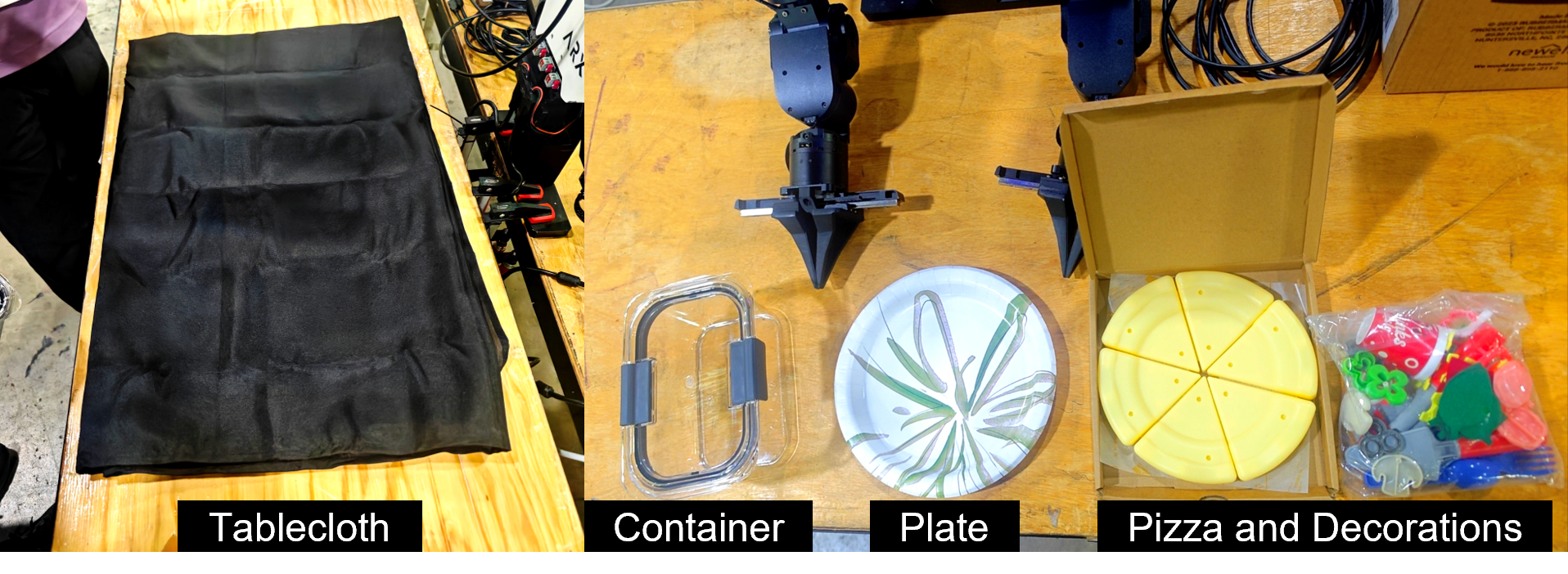}
    \caption{Props and objects provided by the organizers for the Table Service Track.}
\end{figure}

\section{Our Solutions}
\paragraph{Insights from challenge.} Our solution was shaped by a trade-off among competition rules, task characteristics, and current technological capabilities. In particular, the subtasks differ greatly in nature—ranging from long-horizon deformable manipulation (tablecloth unfolding) to force-sensitive precision operations (lid alignment), coordinated locking, and standard pick-and-place (pizza placement). Such diversity makes it difficult for a single approach to handle all cases effectively. Therefore, we adopted a hybrid strategy: applying imitation learning only to the robust pick-and-place task, while relying on remote teleoperation for the more variable or delicate subtasks.
%-------------------------------------------------------------------------
\subsection{In-person Teleoperation}
In-person teleoperation directly connects the operator’s control device with the robot, while allowing the operator to observe the environment on-site.  
To enable \emph{Place the pizza} to be performed by an imitation-learning-based policy, we used in-person teleoperation to collect high-quality demonstrations with minimal latency.  

Specifically, we employed a Meta Quest 3S headset to establish a VR-based teleoperation interface. 
The handheld controllers were retargeted to the end-effectors (EEs) of the X7s dual arms, providing natural mapping of human hand movements to robot arm actions. 
Gripper states were controlled through the controller buttons, ensuring precise and intuitive grasping.  
This setup allowed us to efficiently gather reliable demonstrations, which were later used to train the imitation learning policy.

%-------------------------------------------------------------------------
\subsection{Remote Teleoperation}
Unlike in-person teleoperation, remote teleoperation does not allow the operator to directly observe the manipulation scene.  
Instead, the operator relies on the robot’s onboard sensors, such as the head-mounted or hand-eye cameras (as shown in Fig.~3), to perceive the environment and guide task execution.  
This mode is more challenging but also closer to real-world service applications.  

In our setup, the \textbf{Meta Quest 3S} headset and controllers were used for pose tracking and input.  
Controller poses were retargeted to the X7s dual-arm end-effectors, while buttons operated the grippers.  
Visual feedback was not streamed into the VR device but displayed on a \textbf{desktop monitor} via camera feeds, while the operator manipulated the robot through the VR controllers.  
The master–slave control system was built on \textbf{ROS 1 Noetic}, with all topics transmitted over a local network supported by a \textbf{WiFi 7 router}.  
RGB streams were sent in \textbf{JPEG-compressed format} and decompressed at the receiving end, reducing bandwidth demands and latency.  
Although slower than in-person teleoperation, this configuration was stable, scalable, and assigned a higher operation coefficient, so it was adopted for most subtasks in the competition.

% -------------------------------------------------------------------------
\subsection{Autonomous Policy}
\paragraph{Adopt ACT~\cite{zhao2023learning} for Bimanual Pick-and-place.}
We employed the \textbf{Action Chunking Transformer (ACT)} framework as the policy backbone, due to its effectiveness in stable manipulation.  
Unlike the standard setup with four camera inputs, we adopted a \textbf{three-camera configuration}: two wrist-mounted (hand-eye) cameras and one head-mounted camera.  
This provided complementary egocentric and global perspectives while reducing data bandwidth and redundancy.  

The policy directly predicted robot joint commands instead of task-space poses.  
Each X7s arm had \textbf{7 joints} plus a \textbf{1-DoF gripper}, resulting in a total of \textbf{16 degrees of freedom} for the dual-arm system.  
By outputting coherent joint-level trajectories, ACT enabled smooth coordination between the two arms and reliable gripper control.

\begin{figure}[h]
  \centering
  \includegraphics[width=\columnwidth]{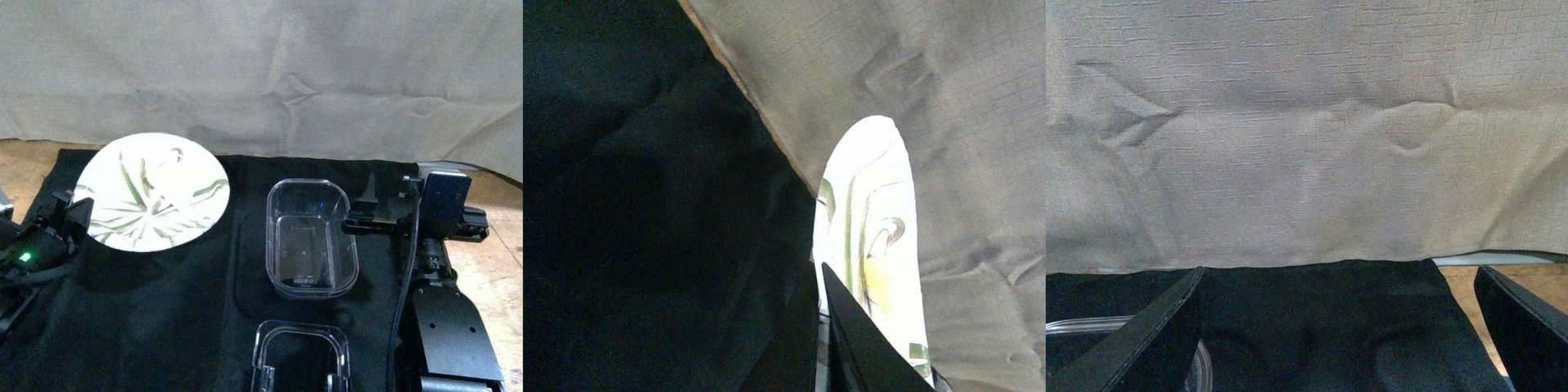}
  % \vspace{-20pt}
  \caption{Visualization of built-in sensor observations during remote teleoperation. From left to right: head-mounted camera, left wrist camera, and right wrist camera.}
\end{figure}

\paragraph{Data Post-processing: Static Frame Trimming.}
During data collection, we observed that demonstration trajectories often contained long static prefixes at the beginning, as the operator prepared for motion.  
These redundant frames introduced bias and degraded policy learning, since the model was forced to imitate stationary states instead of meaningful actions.  

To address this issue, we developed a \textbf{post-processing script} that automatically detects and trims static prefixes based on the motion of both end-effectors (EEF).  
The script monitors frame-to-frame displacements of left and right EEF positions, and removes the initial segment if the displacement stays below a threshold for several consecutive frames.  
The trimmed dataset is then re-saved into clean \texttt{HDF5} files, ensuring that each trajectory starts directly from meaningful motion.  
Additionally, the tool generates plots of EEF trajectories before and after trimming, allowing for quick verification.  
This preprocessing step significantly improved data quality and helped stabilize policy training.

\begin{figure}[h]
  \centering
  \includegraphics[width=\columnwidth]{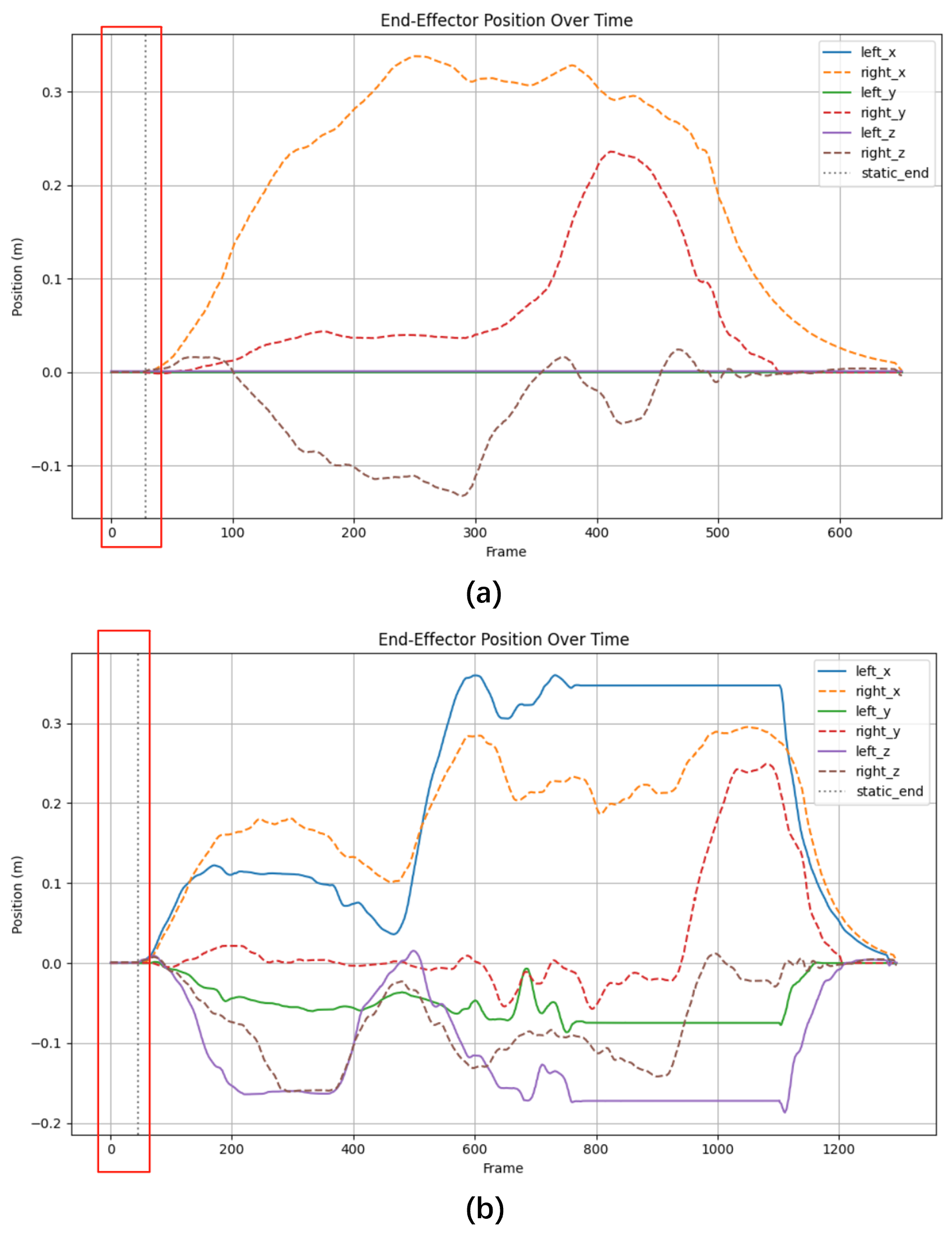}
  % \vspace{-20pt}
  \caption{Examples of data preprocessing by static-frame trimming for unimanual (a) or bimanual (b) demonstrations. The red box highlights end-effector (EE) positions with long idle prefixes that are automatically trimmed.}
\end{figure}

\paragraph{Training Scheme.}
We trained the ACT policy on \textbf{100} in-person teleoperated demonstrations of the pizza pick-and-place task, where the initial poses of plate, container, and pizza were randomized.   
Training used behavior cloning with Adam ($lr{=}1{\times}10^{-5}$), a chunk size of 30, hidden dimension 512, feedforward dimension 3200, batch size 4, and 8000 epochs, with a KL weight of 10 for regularization.

\subsection{Performance}
In the 30-minute final session, our team successfully completed nine task rounds, achieving full scores on all subtasks except for the final round. As shown in Table 1, the completion time of each task varied significantly across rounds: for example, Task 1 (tablecloth unfolding) ranged from less than one minute to over two minutes, depending on the complexity of the folded state; Task 2 (container opening) fluctuated between 10 and 55 seconds, influenced by whether the locks disengaged smoothly; and Task 3 (pizza packing) exhibited the greatest variability due to alignment and locking, occasionally extending beyond 90 seconds. Despite these instabilities, the overall round time averaged about 3 minutes and 20 seconds, enabling us to achieve both high reliability and competitive efficiency under the strict time limit.

\begin{table}[]
\begin{center} 
\caption{Official record for MilkDragon Team's performance.}
\begin{tabular}{@{}lcccc@{}}
\toprule
\textbf{Round} &   & \textbf{Task 1\#} &\textbf{ Task 2\#} & \textbf{Task 3\#} \\ 
\midrule
 & $\alpha$ & 1 & 1 & 1 \\
Round 1\# & $\beta$ & 1'47'' & 0'13'' & 1'56''  \\
 & Score & 5 & 5 & 5  \\
\midrule
 & $\alpha$ & 1 & 1 & 1 \\
Round 2\# & $\beta$ & 1'24'' & 0'40'' & 1'15'' \\
 & Score & 5 & 5 & 5  \\
\midrule
 & $\alpha$ & 1 & 1 & 1  \\
Round 3\# & $\beta$ & 1'03'' & 0'56'' & 1'32''  \\
 & Score & 5 & 5 & 5  \\
\midrule
 & $\alpha$ & 1 & 1 & 1  \\
Round 4\# & $\beta$ & 2'18'' & 0'27'' & 0'30''  \\
 & Score & 5 & 5 & 5  \\
\midrule
 & $\alpha$ & 1 & 1 & 1  \\
Round 5\# & $\beta$ & 2'10'' & 0'55'' & 0'52''  \\
 & Score & 5 & 5 & 5  \\
\midrule
 & $\alpha$ & 1 & 1 & 1  \\
Round 6\# & $\beta$ & 0'58'' & 0'22'' & 0'59'' \\
 & Score & 5 & 5 & 5  \\
\midrule
 & $\alpha$ & 1 & 1 & 1  \\
Round 7\# & $\beta$ & 1'43'' & 0'16'' & 1'24''  \\
 & Score & 5 & 5 & 5 \\
\midrule
 & $\alpha$ & 1 & 1 & 1 \\
Round 8\# & $\beta$ & 1'24'' & 0'28'' & 1'42'' \\
 & Score & 5 & 5 & 5 \\
\midrule
 & $\alpha$ & 1 & 1 & 1  \\
Round 9\# & $\beta$ & 1'24'' & 0'10'' & 0  \\
 & Score & 5 & 5 & 1  \\
\bottomrule
\end{tabular}
\end{center}
\end{table}
\section{Discussion}
\label{sec:Discussion} Our solution demonstrates that combining VR-based teleoperation with imitation learning is an effective strategy for complex bimanual manipulation tasks, balancing robustness and efficiency under competition constraints. Still, challenges remain: policies require many demonstrations to reach high success rates, hardware support for scalable data collection is limited, and the field lacks standardized simulation benchmarks for fair evaluation. Addressing these gaps will be key to pushing bimanual manipulation toward real-world deployment.

\newpage
\section*{Acknowledgements}
We would like to thank the organizers of the ICRA 2025 WBCD Challenge: Zhuo Xu, Dennis Gu and Peter Yu, for their on-site support in coordinating hardware materials and clarifying competition details for our team. We also thank the final-round judges, Di Huang and Yuanzhe Dong, for their efforts. Our appreciation extends to Josh Zhang, Nina Ma and Yongzhen Zou from ARX Robotics for their hardware support during the competition and for sponsoring the championship prize.

{
    \small
    \bibliographystyle{ieeenat_fullname}
    \bibliography{main}
}


\end{document}